\documentclass[preprint,12pt]{elsarticle}

\usepackage{caption}
\usepackage{amssymb}
\usepackage{lscape}
\usepackage{tabularx}
\usepackage{booktabs}
\usepackage{graphicx}
\usepackage{amssymb}
\usepackage{subfigure}
\usepackage{array}
\usepackage{lscape}
\usepackage{dcolumn,longtable,hhline,colortbl}
\usepackage[rotateright]{rotating}
\usepackage{multirow}
\usepackage[section] {placeins}
\usepackage{amssymb}
\usepackage{amssymb}
\usepackage{tabularx}
\usepackage{booktabs}
\usepackage{subfigure}
\usepackage{amsmath}
\usepackage{epsfig}
\usepackage{longtable}
\usepackage{lscape}
\usepackage{epsf}
\usepackage{multirow}

\usepackage{slashbox}

\usepackage{multirow}

\usepackage{graphicx}
\usepackage{epstopdf}

\usepackage{amssymb}

\journal{Expert Systems with Applications}

\begin{document}
\begin{frontmatter}



\title{Defuzzify firstly or finally: Dose it matter in fuzzy DEMATEL under uncertain environment?}


\author[address1]{Yunpeng Li}
\author[address1]{Ya Li}
\author[address1]{Jie Liu}
\author[address1,address3]{Yong Deng\corref{label1}}

\cortext[label1]{Corresponding author: Yong Deng, School of
Computer and Information Science, Southwest University, Chongqing,
400715, China. Email address: ydeng@swu.edu.cn,
prof.deng@hotmail.com. Tel/Fax:(86-023)68254555.}

\address[address1]{School of Computer and Information Science, Southwest University, Chongqing, 400715, China}
\address[address3]{School of Engineering, Vanderbilt University, Nashville, TN, 37235, USA}

\begin{abstract}

Decision-Making Trial and Evaluation Laboratory (DEMATEL) method is widely used in many real applications. With the desirable property of efficient handling with the uncertain information in decision making, the fuzzy DEMATEL is heavily studied. Recently, Dytczak and Ginda suggested to defuzzify the fuzzy numbers firstly and then use the classical DEMATEL to obtain the final result. In this short paper, we show that it is not reasonable in some situations. The results of defuzzification at the first step are not coincide with the results of defuzzification at the final step.It seems that the alternative is to defuzzification in the final step in fuzzy DEMATEL.

\end{abstract}

\begin{keyword}
Fuzzy sets; DEMATEL; Defuzzify;Deasion making
\end{keyword}
\end{frontmatter}



\section{Introduction}

Decision making is a very important process in many intelligent systems. A number of methodologies are developed to handle decision making. One of the widely used methods is Decision-Making Trail and Evaluation Laboratory (DEMATEL), originally presented by Fontela and Gabus \cite{fontela1976dematel}. Due to its efficiency in modelling of the structure of the\emph{ cause-effect} relationships (directed or influences) between different components in the complex systems, the DEMATEL has been widely used in many fields \cite{tseng2009application,zhou2011identifying,wang2012brand,fan2012identifying,hsu2012evaluation,wang2012dematel,lee2013revised}. In order to address the modelling of uncertain information, the DEMATEL is generalized with fuzzy sets theory to handle decision making under uncertain environment \cite{tseng2010assessment,chou2012evaluating,buyukozkan2012novel,wu2012segmenting,hiete2012trapezoidal,zhou2013development}.

Recently, some researches argue that a concept of explicit fuzzy data processing in DEMATEL needs a thorough rethink \cite{dytczak2013explicit}. The authors suggest to transform the fuzzy evaluations into crisp data prior to the actual processing of data using DEMATEL. In this short paper, we use a numerical example to show that there is the difference in the results between the existing fuzzy DEMATEL and the proposed method by Dytczak and Ginda \cite{dytczak2013explicit}, which will have the effect on the final decision making. In other words, it does matter in the decision making under uncertain environment based on fuzzy DEMATEL that in which step to transform the fuzzy numbers, in the first step or in the final step.

The remain of this paper is organized as follows. Section 2 presents some preliminaries, including the brief introduction of the method proposed by Dytczak and Ginda \cite{dytczak2013explicit}. We use a numeral example to show the shortcoming of Dytczak and Ginda's method in section 3. In section 4 we argue that, either in the aspect of qualitative or in the aspect of quantitative, it is not a real alternative to transform fuzzy numbers to corresponding crisp data prior to the processing of data using DEMATEL procedure. Section 5 ends this paper with a brief conclusion.

\section{Problem Description}\label{sec:fuzzy-DEMATEL}
The distinction of these fuzzy-DEMATEL methods focuses on the choice of fuzzy set, which includes triangular fuzzy numbers, trapezoidal fuzzy numbers, L-R fuzzy numbers \textit{et.c.}\cite{opricovic2003defuzzification}. In this paper, the same fuzzy-DEMATEL method with \cite{dytczak2013explicit} is applied.

The fuzzy evaluations used in this paper are based on triangular fuzzy numbers(TFNs). TFN ${X}$ is composed of three crisp numbers: lower limit($x^{(l)}$), medium value($x^{(m)}$), and upper limit($x^{(u)}$). TFN is widely used in fuzzy-DEMATEL methods\cite{hiete2012trapezoidal}. Detailed analysis of TFN would not be included for it is not the focus of this paper. Further description of TFN and fuzzy set theory can be found in \cite{tseng2009causal}.

When the fuzzy evaluations are acquired, the crisp evaluations are obtained by applying defuzzification to the fuzzy evaluations. However, the defuzzification operation cannot be defined as a single universal process\cite{opricovic2003defuzzification}. In this paper, the TFNs are defuzzified with Eq.(\ref{equ:defuzzy}) in correspondence with the discussion paper\cite{dytczak2013explicit}.

\begin{equation}\label{equ:defuzzy}
x = x^{(m)}+\frac{x^{(u)^2}+2\cdot x^{(m)}\cdot(x^{(l)}-x^{(u)})-x^{(l)^2}}{3\cdot(x^{(u)}-x^{(l)})}\\
  = \frac{x^{(l)}+x^{(m)}+x^{(u)}}{3}
\end{equation}

The crisp evaluations and fuzzy evaluations are then processed by crisp DEMATEL and fuzzy-DEMATEL methods, respectively. Fuzzy-DEMATEL treats the lower limit, middle value and upper limit separately with crisp DEMATEL method. The numeric result of fuzzy evaluations is defuzzified at the end of the fuzzy-DEMATEL process with Eq.(\ref{equ:defuzzy}). Detailed description of the process can be found in the discussion paper\cite{dytczak2013explicit}.

\section{From the view of qualitative}\label{sec:counter_example}
We have discovered several cases in which crisp DEMATEL and fuzzy-DEMATEL derive different ranks of objects. Following is one of the counter examples.

\begin{figure}[!ht]
\begin{center}
  \includegraphics[scale=1.2]{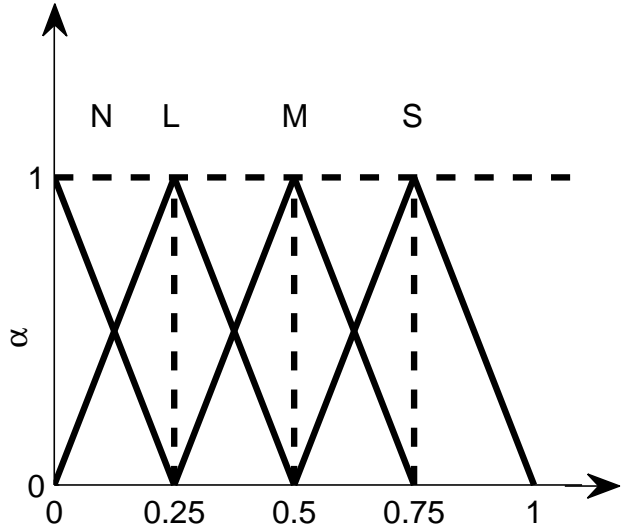}
\caption{Fuzzy DEMATEL scale levels}
\label{graph:scale}       
\end{center}
\end{figure}

The counter example is composed of 5 objects: A,B,C,D,E. Crisp scale is composed of 4 levels rank from 0 (lack of influence) to 3 (extreme influence). Fuzzy scale have corresponding levels to crisp scale levels (N,L,H,S). The fuzzy set of each level is shown in Fig.\ref{graph:scale}.Both crisp and fuzzy scales are identical to the proposed discussion paper\cite{dytczak2013explicit}.

\begin{figure}[!ht]
\begin{center}
  \includegraphics[scale=0.8]{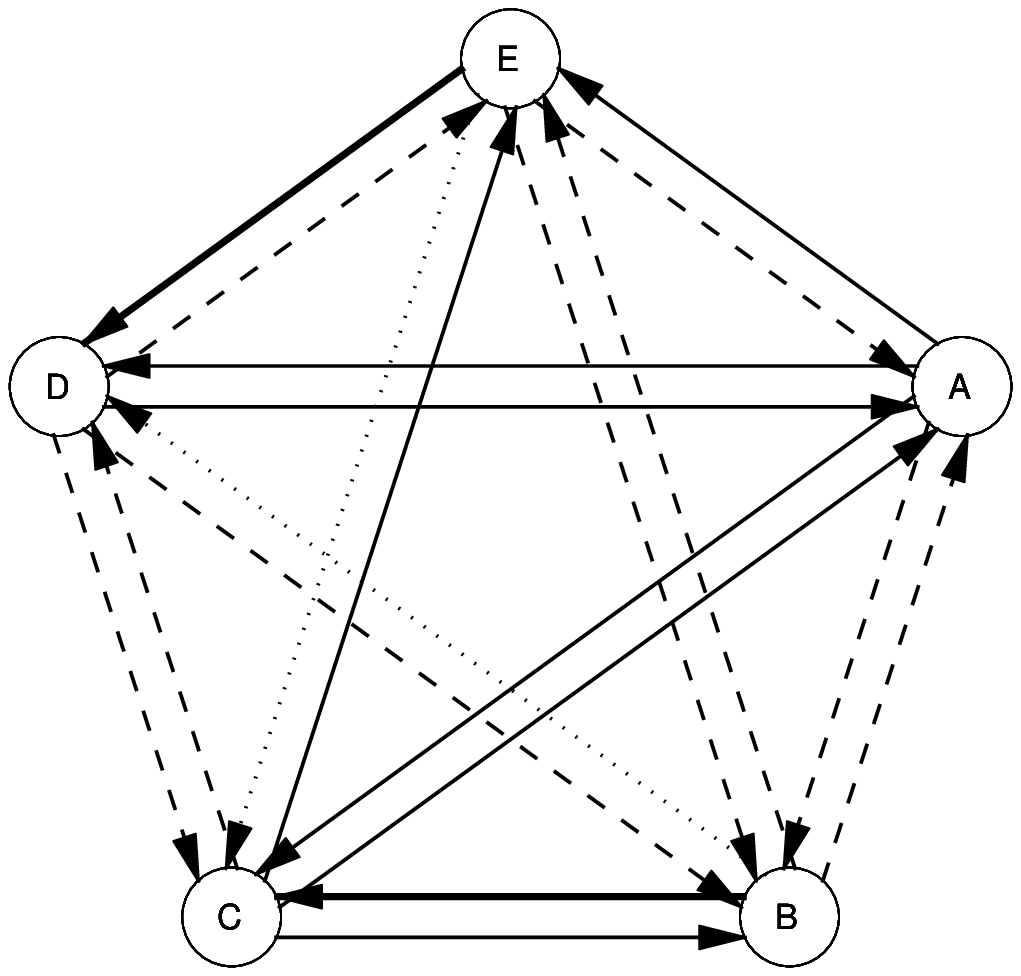}
\caption{Structures of direct influence of the counter example}
\label{graph:structure}       
\end{center}
\end{figure}

Direct influence of objects is evaluated by a single expert. The structure of influences between factors in the counter example is shown in Fig.\ref{graph:structure}. Line patterns of arcs correspond to different direct influence levels:

1. Dotted line pattern corresponds to the 0/N scale level.

2. Dashed line pattern expresses the 1/L scale level.

3. Solid line pattern corresponds to the 2/H scale level.

4. Bold line pattern denotes the 3/S scale level.

\begin{figure}[!ht]
\begin{center}
  \includegraphics[scale=0.8]{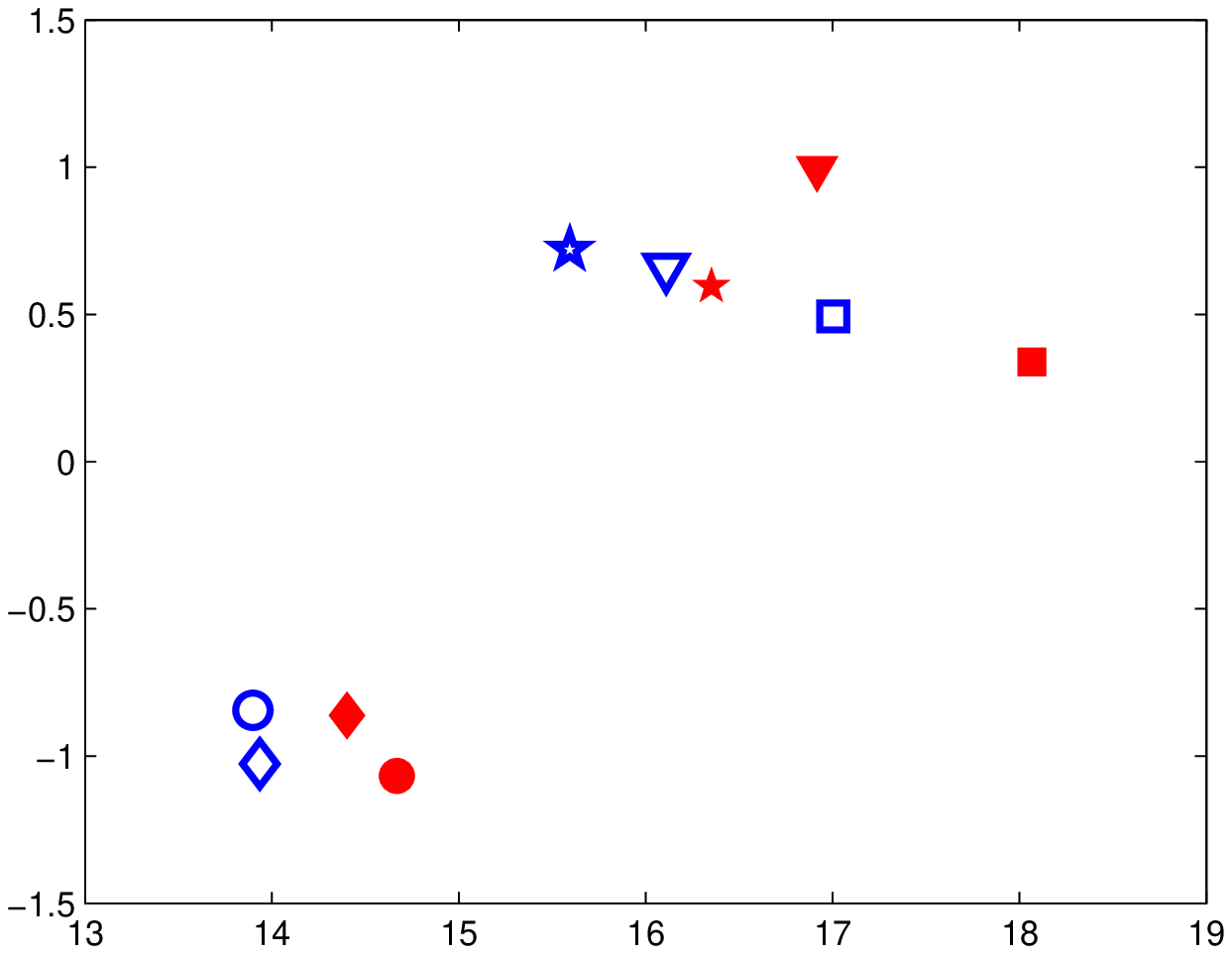}
\caption{Crisp (red) and fuzzy (blue) DEMATEL results}
\label{graph:CE}       
\end{center}
\end{figure}

We applied the crisp DEMATEL method as well as fuzzy-DEMATEL method to our counter example. Result is shown in Fig.\ref{graph:CE}. Red markers denote to crisp DEMATEL result, and blue markers denote to fuzzy-DEMATEL result.

As can be seen from Fig.\ref{graph:CE}, the ranks of R+C and R-C are different in crisp model and fuzzy model. The detailed results are provided in Table.\ref{table:R-C}

\begin{table}\scriptsize
\caption{Results of Crisp DEMATEL and Fuzzy DEMATEL}
\begin{tabular}{ccccccccccccc}
\hline
\multirow{2}{*}{Object} & \multicolumn{2}{c}{R} & \multicolumn{2}{c}{C} & \multicolumn{2}{c}{R+C} & \multicolumn{2}{c}{R-C} & \multicolumn{2}{c}{R+C Rank} & \multicolumn{2}{c}{R-C Rank}\\
\cline{2-13}
& Cr. & Fu. & Cr. & Fu. & Cr. & Fu. & Cr. & Fu. & Cr. & Fu. & Cr. & Fu.\\
\cline{1-13}
A & 6.801 & 6.527 & 7.868 & 7.372 & 14.67 & 13.90 & -1.067 & -0.845 & 4 & 5 & 5 & 4\\
B & 9.202 & 8.749 & 8.864 & 8.256 & 18.07 & 17.01 & 0.339 & 0.493 & 1 & 1 & 3 & 3\\
C & 6.770 & 6.454 & 7.632 & 7.481 & 14.40 & 13.94 & -0.862 & -1.026 & 5 & 4 & 4 & 5\\
D & 8.956 & 8.384 & 7.960 & 7.725 & 16.92 & 16.11 & 0.996 & 0.658 & 2 & 2 & 1 & 2\\
E & 8.474 & 8.157 & 7.878 & 7.437 & 16.35 & 15.59 & 0.595 & 0.720 & 3 & 3 & 2 & 1\\
\hline
\end{tabular}
\label{table:R-C}
\end{table}

It is obviously shown that the R+C  and R-C of objects altered dramatically. A and C have their ranks exchanged, both R+C and R-C. At the same time, D and E have their R-C ranks exchanged. Technically, the result can be easily justified: Each object have decreased values of R and C when evaluated with fuzzy rather than crisp DEMATEL method. However, such decrease are not in the same amount. Yet the sum of R-C for all objects is always fixed to 0. So the R-C column have to change in different directions, positive and negative. This change may lead to alternation in ranking if two objects, close enough, change simultaneously to different directions. In this case, such pairs are object A\&C and D\&E. In short,from the view of qualitative, the ranking results of crisp model proposed by Dytczak and Ginda \cite{dytczak2013explicit} are not coincide with the classical Fuzzy DEMATEL.

\section{From the view of quantitative}\label{quantitative}

It is inevitable to cause information distortion in the process of defuzzification. The main reason is that a single value is used to represent a fuzzy numbers, which is, actually a set of values. For a designer or manager of a decision support system, it is more reasonable that the information distortion in the system is as less as possible. The fuzzy model and the crisp model proposed by Dytczak and Ginda \cite{dytczak2013explicit} is compared from the view of quantitative in this section.

\begin{landscape}
\begin{figure}[!ht]
\begin{center}
  \includegraphics[scale=0.9]{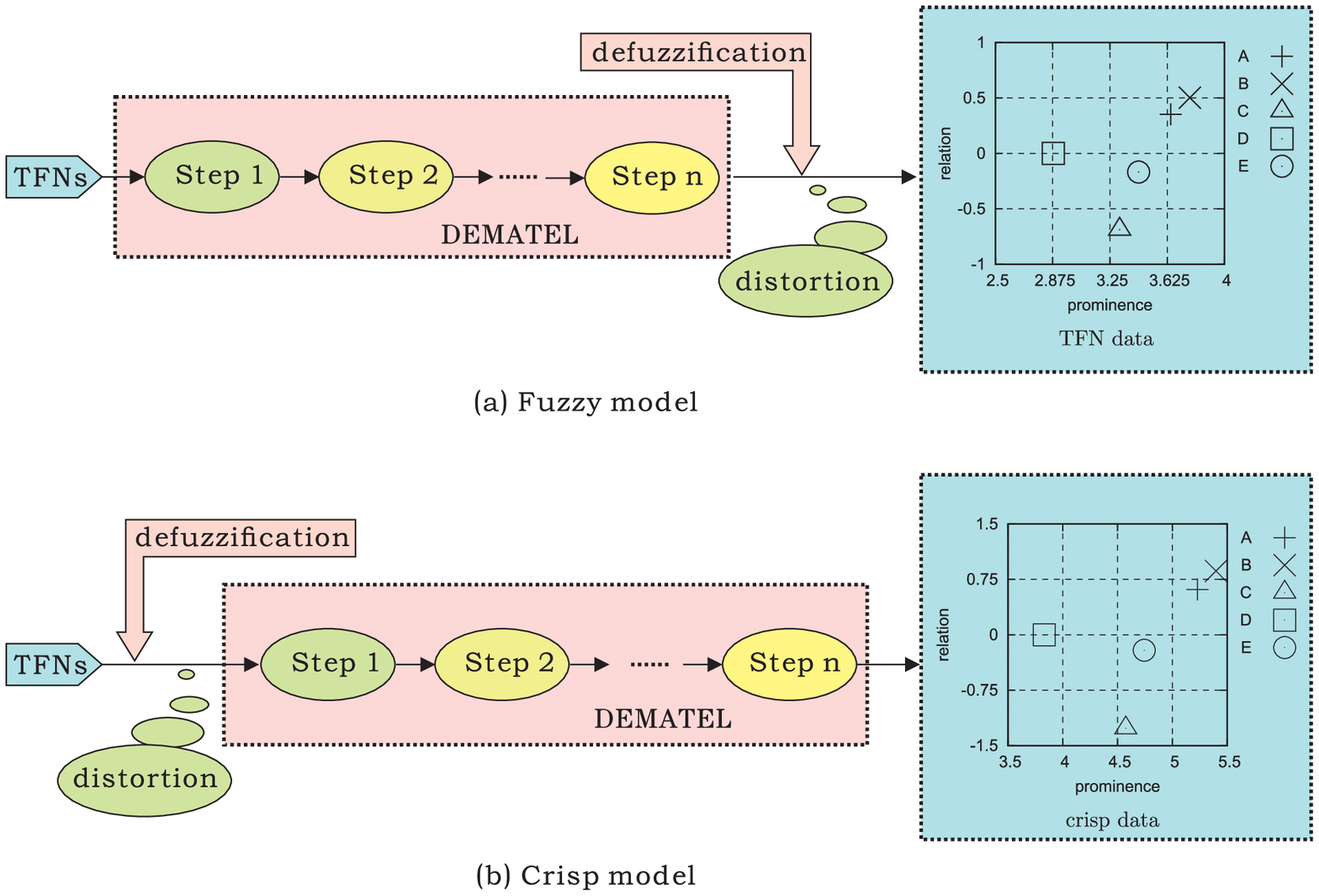}
\caption{Comparation between fuzzy DEMATEL model and crisp DEMATEL model}
\label{Comparation}
\end{center}
\end{figure}
\end{landscape}

As shown in Fig. \ref{Comparation}, the inputs of both models are TFNs. In the fuzzy DEMATEL  model, the defuzzification is arranged in the last step, while the  defuzzification in crisp DEMATEL model is set in the first step. Comparing Fig. \ref{Comparation}(a) to (b), we should remark that the defuzzification in difference step has result in alternation of values of relation and prominence. The ranges of relation in crisp model is -1.4 to 0.9, while in fuzzy model is -0.75 to 0.6. The minimum value of relation in crisp model is nearly twice lower than that in fuzzy model. It is similar as in prominence. The range of relation and prominence in fuzzy model is less than that in crisp model. The main reason for this is due to the uncertain information in fuzzy model is maintained during the whole process of DEMATEL until the last step. From Fig. \ref{Comparation} we can see that the information distortion arise earlier in crisp model. For a decision maker,he or she may face a question which model is more preferable to make a confidential decision. If they want to dutifully maintain the uncertainty of data under uncertain environment, they would prefer fuzzy mode due to the less information distortion.

\section{Conclusion}

It should be pointed out that there are many open issues in fuzzy DEMATEL such as the higher computational complexity than classical DEMATEL. However, the method to defuzzify fuzzy numbers at the first step to use DEMTATEL is not a real alternative. We use a numerical example to show that, even in the situation we takes only the qulative into consideration, the results of defuzzify firstly are not coincide with those of defuzzify finally. From the point of reliable degree, defuzzication in final step is more reasonable since the information distortion in this manner is better than defuzzification in the first step. To our opinions, with the development of computer science, the cost on computational complexity is deserved for a more confidential decision making in fuzzy DEMATEL.

\section*{Acknowledgment}
The work is partially supported by National Natural Science Foundation of China, Grant No. 61174022, Chongqing Natural Science Foundation for Distinguished Young Scientists,
Grant No. CSCT, 2010BA2003, Program for New Century Excellent Talents in University, Grant No.NCET-08-0345, National High Technology Research and Development Program of China (863 Program) (No.
2013AA013801).








\bibliographystyle{elsarticle-num}
\bibliography{Ya}







\end{document}